\documentclass[runningheads]{llncs}

\usepackage{eccv}

\usepackage{eccvabbrv}

\usepackage{graphicx}
\usepackage{booktabs}
\usepackage{xcolor,colortbl}
\usepackage{comment}
\usepackage[accsupp]{axessibility}  
\definecolor{lightgray}{rgb}{0.88, 0.88, 0.88}

\usepackage{hyperref}

\usepackage{orcidlink}
\usepackage{graphicx}
\usepackage{amsmath}
\usepackage{amssymb}
\usepackage{booktabs}
\usepackage{multirow}
\usepackage{tabularx}
\usepackage{bm}
\usepackage{wrapfig}
\newcolumntype{g}{>{\columncolor{Gray}}c}
\usepackage{physics}
\newcolumntype{C}{>{\centering\arraybackslash}X} 

\begin{document}

\title{EMAG: Ego-motion Aware and Generalizable \\ 2D Hand Forecasting from Egocentric Videos}

\titlerunning{EMAG}

\author{Masashi Hatano\inst{1}\orcidlink{0009-0002-7090-6564} \and
Ryo Hachiuma\inst{2}\orcidlink{0000-0001-8274-3710} \and
Hideo Saito\inst{1}\orcidlink{0000-0002-2421-9862}}

\authorrunning{M.~Hatano et al.}

\institute{
Keio University \and
NVIDIA
}

\maketitle

\begin{abstract}
Predicting future human behavior from egocentric videos is a challenging but critical task for human intention understanding.
Existing methods for forecasting 2D hand positions rely on visual representations and mainly focus on hand-object interactions.
In this paper, we investigate the hand forecasting task and tackle two significant issues that persist in the existing methods: (1) 2D hand positions in future frames are severely affected by ego-motions in egocentric videos; (2) prediction based on visual information tends to overfit to background or scene textures, posing a challenge for generalization on novel scenes or human behaviors.
To solve the aforementioned problems, we propose EMAG, an ego-motion-aware and generalizable 2D hand forecasting method.
In response to the first problem, we propose a method that considers ego-motion, represented by a sequence of homography matrices of two consecutive frames.
We further leverage modalities such as optical flow, trajectories of hands and interacting objects, and ego-motions, thereby alleviating the second issue.
Extensive experiments on two large-scale egocentric video datasets, Ego4D and EPIC-Kitchens 55, verify the effectiveness of the proposed method.
In particular, our model outperforms prior methods by $1.7$\% and $7.0$\% on intra and cross-dataset evaluations, respectively.
Project Page: \url{https://masashi-hatano.github.io/EMAG/}

    
\end{abstract}

\section{Introduction}
\label{sec:intro}

With the emergence of wearable devices such as smart glasses and intelligent helmets, there has been growing interest in the analysis of egocentric videos.
In recent years, large-scale egocentric vision datasets such as EPIC-Kichtens~\cite{epic-55, epic-100} and Ego4D~\cite{ego4d} have been introduced to catalyze the next era of research in first-person perception and provide a diverse range of tasks for investigation, including action recognition~\cite{Wang_2021_ICCV, Plizzari_2022_CVPR, Gong_2023_CVPR}, human body pose estimation~\cite{Wang_2021_ICCV_mpi, Wang_2023_CVPR, Li_2023_CVPR}, audio-visual understanding~\cite{Huang_2023_CVPR, Ryan_2023_CVPR}, action anticipation~\cite{rulstm, srl}, and natural language queries~\cite{spotem}.

Future forecasting is one of the major categories, including the anticipation of the camera wearer's future actions and the prediction of human movements.
This capability has immediate applications in AR/VR~\cite{wilmottismar2022, xu2019ismar} and human-robot interactions~\cite{whitney2016icra, quintero2018iros} as both fields benefit from understanding the camera wearer's actions or behaviors. 
Among the tasks in future forecasting, hand forecasting has been recognized as particularly challenging due to severe ego-motion, which affects the 2D hand positions in future frames.

\begin{wrapfigure}{r}{0.5\linewidth}
\begin{center}
\vspace{-3em}
   \includegraphics[width=\linewidth]{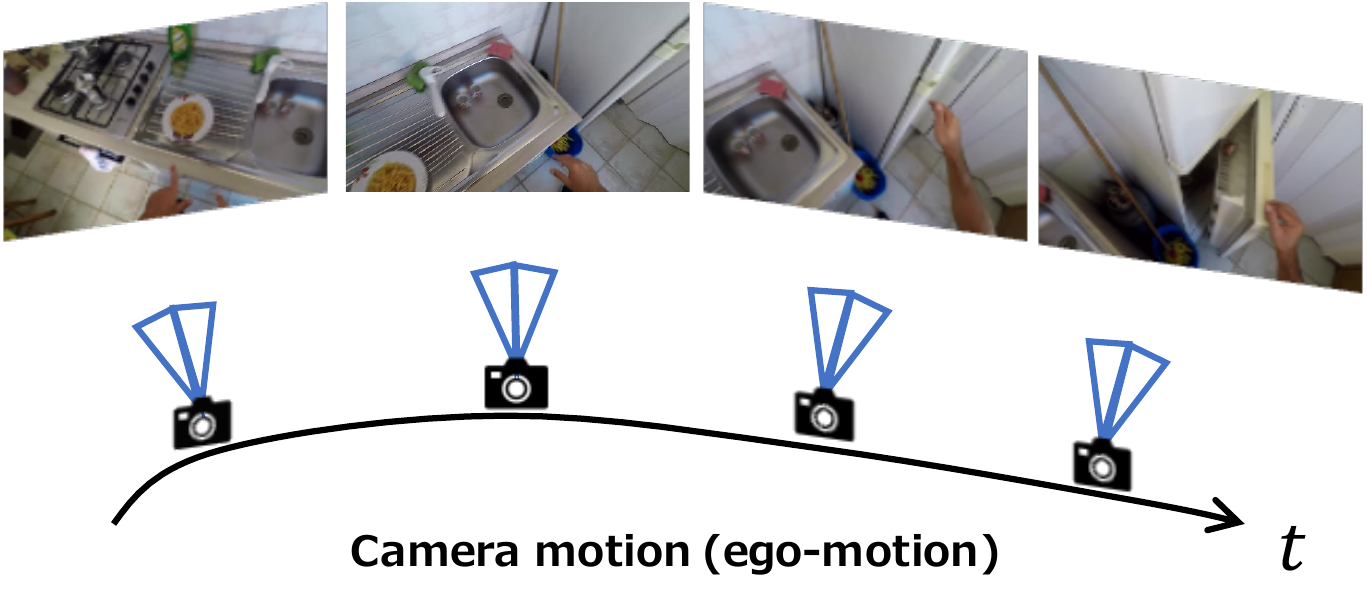}
\end{center}
\vspace{-2em}
   \caption{The presence of ego-motion in first-person videos significantly affects the dynamic movement of the camera position. Since the camera is part of the wearer's body, a variety of views can be captured even in a short period of time.}
\label{fig: egomotion}
\vspace{-2em}
\end{wrapfigure}

Recent 2D egocentric hand forecasting approaches~\cite{FHOI, hoi-forecast, ego4d} leverage visual feature representations extracted from input RGB videos using 2D or 3D Convolutional Neural Networks (CNNs) for the hand forecasting task.
For example, the method proposed in the Ego4D dataset~\cite{ego4d} uses a simple I3D network~\cite{I3D} and regresses the future 2D hand coordinates. 
Meanwhile, the Object Centric Transformer (OCT)~\cite{hoi-forecast} is a method that jointly predicts hand motions and object contact points from RGB video features extracted with BNInception~\cite{BNInception} and the hand/object bounding boxes.

Although the 2D hand forecasting task has been widely studied, two critical issues still remain in the previous works: the accuracy and generalization performance against unseen data, both of which are crucial for practical scenarios.
First, the 2D hand position in future frames is heavily influenced by the head motion of the camera wearer, also known as \textit{ego-motion}.
As illustrated in \cref{fig: egomotion}, body and head motions cause frequent view changes even in a short period of time, yet the previous approaches have not explicitly considered ego-motion for predicting 2D hand positions.
Second, the performance of RGB-based prediction approaches significantly drops when the video feature distribution (\ie domain) diverges from that of the training set~\cite{Kim_2021_ICCV, weinzaepfel2021mimetics}. 
This performance drop is crucial for the 2D egocentric hand forecasting task since the camera is not situated at a fixed location.
For instance, performance may vary if the egocentric videos are captured in different textured environments (\eg, outdoor vs. indoor), or if the wearer performs different actions from the training.

This work proposes \textit{EMAG}, an ego-motion-aware and generalizable 2D hand forecasting method.
This approach capitalizes on the incorporation of ego-motion information to enhance the accuracy of the hand forecasting task.
Additionally, we employ multiple modalities to mitigate susceptibility to overfitting in backgrounds or scene textures. 
We aim to achieve more robust predictions in settings where camera wearers engage in a diverse range of tasks such as cooking and gardening.

To address the first challenge, we propose leveraging a sequence of homography matrices as ego-motion and anticipating them on future frames.
Given that hand positions in future frames are affected by future ego-motion, explicitly forecasting ego-motion as an auxiliary task enhances the accuracy of predicting future hand positions, particularly in egocentric videos where head motions occur frequently.

To alleviate the second issue, instead of primarily relying on visual features for estimating 2D hand positions, we leverage modalities such as optical flow, hand/object positions, and ego-motion information, using hand motions as the primary features for hand forecasting.
This approach reduces reliance on appearance-based features, as these modalities are free from appearance-based biases~\cite{Plizzari_2022_CVPR}. 
Consequently, the model's generalizability is enhanced, ensuring robust performance even when distribution gaps exist between the training and test data.

We extensively evaluate the proposed method on two large-scale egocentric datasets, Ego4D~\cite{ego4d} and EPIC-Kitchens 55~\cite{epic-55}. 
The performance of the proposed method, along with that of previous state-of-the-art forecasting approaches, is assessed under two settings: the intra-dataset setting and the cross-dataset setting.
In the cross-dataset setting, the model is evaluated on a different dataset from training to verify the generalization performance against unseen scenes or actions. 
As a result, our method outperforms the previous approaches in both two settings ($1.7\%$ and $7.0\%$ improvement with intra-dataset and cross-dataset settings, respectively). 
Moreover, we conduct various ablation studies on the proposed input modalities and loss components.

In summary, our contributions are as follows:
\begin{itemize}
    \item We are the first to investigate the potential benefits of incorporating ego-motion, which is critical in the 2D hand forecasting task.
    \item We propose a simple but effective approach, EMAG, that considers ego-motion, represented by a sequence of homography matrices of two consecutive frames.
    In addition, our method utilizes multiple modalities to mitigate overfitting to scene textures.
    \item We conduct extensive experiments on two large-scale egocentric datasets, Ego4D and EPIC-Kitchens 55. The experimental results verify the outperformance of the proposed method over the previous approaches through two different experimental setups: intra-dataset and cross-dataset. Especially, the method shows strong performance with cross-dataset in which the training and test datasets differ.
\end{itemize}
\section{Related Work}
\label{sec:related}

\subsection{Egocentric Video Understanding} 
Video understanding is one of the central tasks in the computer vision field. 
Various video understanding methods are well-established thanks to large-scale datasets~\cite{kinetics-700, ava, sports-1m} collected from internet sources (\eg YouTube).
The videos in these large-scale datasets are mostly captured from an exocentric camera (third-view video), such as a surveillance or a hand-held camera. 

On the other hand, analyzing egocentric video (first-view video) captured by wearable cameras has become an active area of research in recent years~\cite{ohkawa:cvpr23, miaoeccv2022, ego-exo, Xue_2023_CVPR, Price_2022_CVPR, egonerf}. 
Compared with exocentric videos, egocentric videos provide distinct viewpoints of surrounding scenes and actions driven by the camera position holding on the observer.
Therefore, egocentric video analysis can be helpful for various applications, such as AR/VR~\cite{wilmottismar2022, xu2019ismar} or medical image analysis~\cite{birkfellner:tmi2002, fotouhi:tmi2021}.

Multiple large-scale egocentric video datasets ~\cite{ego4d, epic-55, epic-100, egopat, hoi4d} have been proposed in response to the demand for egocentric video analysis.
These datasets have played a pivotal role in advancing research on egocentric video understanding, encompassing tasks such as activity recognition~\cite{Gong_2023_CVPR, Plizzari_2022_CVPR, Wang_2021_ICCV}, human-object interaction~\cite{FHOI, hoi-forecast, Yu_2023_WACV}, action anticipation~\cite{rulstm, srl}, human body pose estimation~\cite{Wang_2021_ICCV_mpi, Wang_2023_CVPR, Li_2023_CVPR}, and audio-visual understanding~\cite{Huang_2023_CVPR, Ryan_2023_CVPR}. 
In this work, we explore one of the challenging tasks in egocentric video analysis, 2D hand forecasting.

\subsection{Hand Forecasting from Egocentric Videos} 
To predict future hand positions, traditional tracking or sequential methods, such as Kalman Filter (KF)~\cite{KF}, Constant Velocity Model (CVM)~\cite{cvm}, and Seq2Seq~\cite{seq2seq}, have been commonly employed for trajectory prediction. 
These methods often rely solely on trajectories of hand positions and do not effectively leverage the context of scenes without visual information, resulting in suboptimal performance. 
To effectively leverage visual information, the baseline method for hand forecasting, which was proposed as a benchmark along with the Ego4D~\cite{ego4d} dataset, utilized I3D~\cite{I3D}, a method that is known for its outstanding performance to extract spatial and temporal information.

Moreover, several studies have focused on hand-object interactions to explore the relationship between meaningful human body movements and future representations.
FHOI~\cite{FHOI} is the first work to incorporate the future trajectory of hands for action anticipation in egocentric videos.
Building upon this, OCT~\cite{hoi-forecast} is an approach that integrates hand-object interactions into the prediction process.

However, neither of these approaches explicitly considers ego-motion, which plays a crucial role in accurately predicting future hand positions in 2D image coordinates, as future hand positions are heavily influenced by future ego-motion.
In contrast to previous works, we explore the potential benefits of integrating ego-motion information to enhance the capability of predicting future hand positions even in the presence of severe ego-motion.

\begin{figure}[t]
\begin{center}
\includegraphics[width=0.95\linewidth]{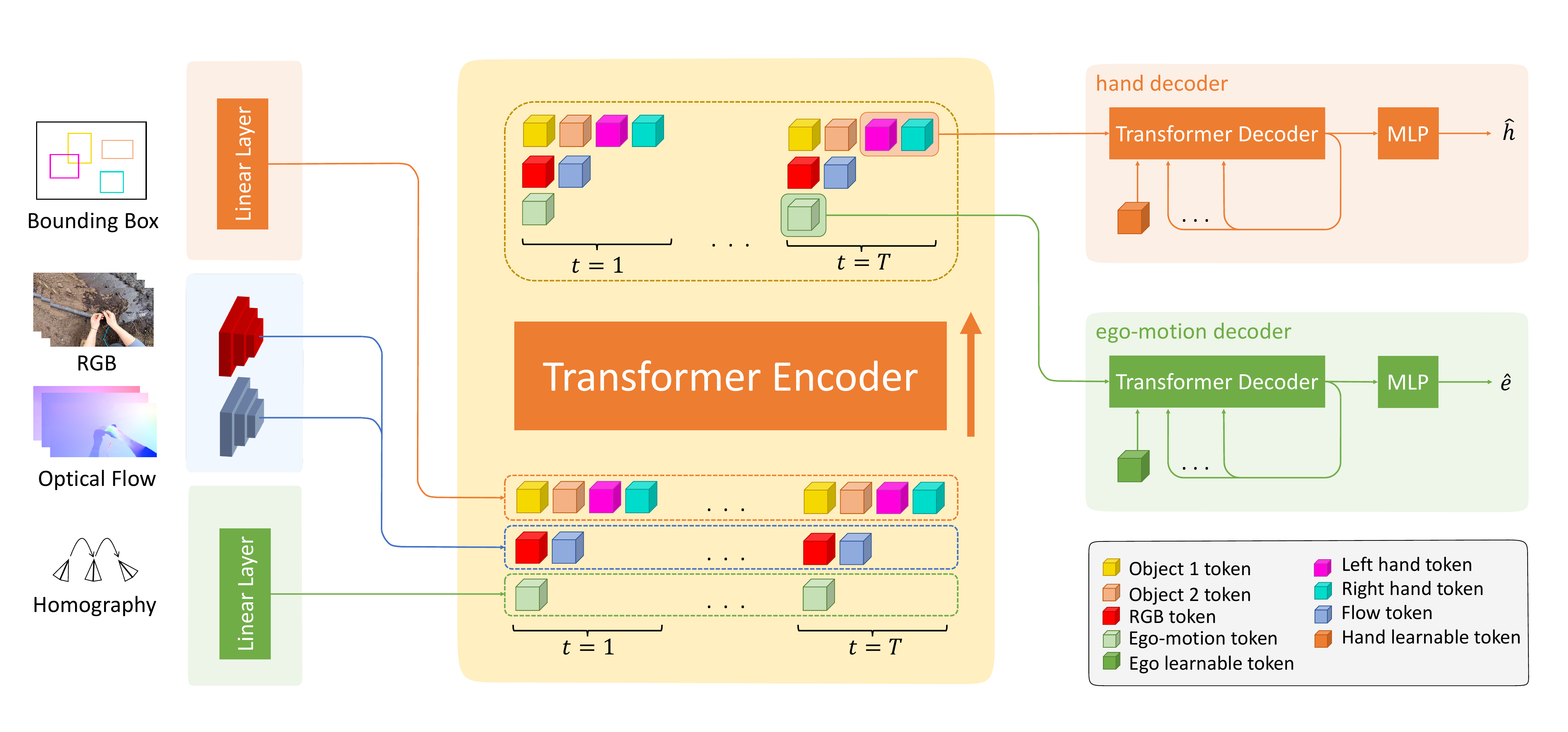}
\end{center}
   \caption{The architecture of the proposed method. Given input egocentric video frames, we pre-process them and obtain multiple modalities, including RGB and optical flow, detected bounding boxes of objects/hands, and homography matrices of adjacent frames.
   We train a single Transformer encoder and two Transformer decoders with MLP heads for hand and ego-motion prediction.}
\label{fig: overview}
\end{figure}
\section{Method}

The proposed architecture is built upon the original Transformer~\cite{transformer}. 
It inputs multiple modalities and predicts future hand positions and ego-motions.
We first introduce the egocentric 2D hand forecasting task (\cref{sec: fhp-task}). 
Then, we introduce our proposed method, including pre-processing (\cref{sec: pp}), an encoder (\cref{sec: enc}), our hand position and ego-motion predictors (\cref{sec: hand-ego}), and our training objective (\cref{sec: to}). 
\cref{fig: overview} provides an overview of our approach.

\subsection{Problem Definition}
\label{sec: fhp-task}
The task is to predict future hand positions of the camera wearer in 2D image coordinates on future frames, followed by the definition on Ego4D~\cite{ego4d}.
Given an input egocentric video $V = \{I^{1}, \ldots, I^{T}\}$ with an observation time length $T$, where $I^{T}$ represents the last observation frame.
Our goal is to predict future hand coordinates $\bm{h} = \{\bm{h}^{T+1}, \ldots, \bm{h}^{T+F}\}$ for the future time horizons $F$.
At each time step $t$, $\bm{h}^{t}$ consists of left/right-hand positions in the 2D image coordinate system on the frame $I^{t}$. 

\subsection{Pre-processing}
\label{sec: pp}
Our proposed method inputs three types of input modalities: trajectory information, global information, and ego-motion information.
We pre-process an input video to obtain these three modalities as follows. 

\noindent\textbf{Trajectory information.} 
Trajectory information consists of the sequential 2D positions of the bounding boxes of hands and objects. 
To obtain bounding boxes for both hands and objects for each frame, we apply an egocentric hand-object detector~\cite{hod}, which detects the left and right hand and objects separately.
We use the following bounding boxes: left hand, right hand, and objects detected with a top-$k$ confidence score.

\noindent\textbf{Global information.} 
Global information consists of RGB frames and optical flow.
The optical flow can be estimated from two consecutive RGB frames via an off-the-shelf optical flow estimator, such as RAFT~\cite{raft} or FlowFormer~\cite{flowformer}.

\noindent\textbf{Ego-motion information.} 
The ego-motion is represented by a sequence of homography matrices, which encapsulate the transformation between consecutive frames.
Generally, a homography between images taken from two distinct viewpoints depends on the intricate 3D arrangement of the captured scene. 
Nonetheless, given the relatively small magnitude of the translation vector connecting consecutive frames in the context of first-person videos, a homography does not depend on the 3D structure of the scene but solely on the rotation between the two viewpoints.

The process of estimating the homography matrix involves two key stages: the identification of matching points between frames and the determination of a homography matrix that minimizes the error. 
The initial step entails identifying matching points, a task facilitated by using previously estimated optical flow, which characterizes the pixel displacement between frames. 
For the second step, we apply the RANSAC algorithm~\cite{ransac}, which is known as a robust iterative algorithm, to estimate the homography parameters.

\subsection{Encoder}
\label{sec: enc}
\noindent\textbf{Tokenization.}
After pre-processing all input modalities, each modality is transformed into a token to be encoded in a single Transformer encoder.
For each detected bounding box (top-left and bottom-right coordinates) at time step $t$, it is transformed into a token $\bm{x}^{t}_{i}$ by a shared linear layer, which maps $\mathbb{R}^{4} \rightarrow \mathbb{R}^{C}$, where $i$ represents either of the left hand, right hand, or objects detected with a top-$k$ confidence score, and $C$ denotes the dimension size of each token.
As for the global information, we use two 2D CNNs to extract the features of each RGB and flow frame and then pool the extracted features in the spatial direction by global average pooling (GAP).
The pooled features are denoted as $\bm{x}^{t}_\text{rgb}$ and $\bm{x}^{t}_\text{flow}$.
Similar to the trajectory information, each homography matrix is transformed into a token $\bm{x}^{t}_{ego}$ by a linear layer, which maps $\mathbb{R}^{9} \rightarrow \mathbb{R}^{C}$.
The $3\times3$ homography matrix is flattened before passing through the linear layer.

\noindent\textbf{Index encoding.}
As there are various tokens in terms of modality type and time, two index encodings, the modal index embedding and time index embedding, are employed.
The learnable position embedding is employed for the modal index embedding.
Also, we adopt the time index encoding, which
replaces the position in the original sinusoid positional encoding~\cite{transformer} with a time index (frame number).

\noindent\textbf{Transformer encoder.}
We use a single Transformer encoder $\mathcal{E}$ to encode multiple input modalities across multiple time steps via self-attention mechanisms:
\begin{equation}
\label{eq: 1}
    \bm{z}^{1}_{m_1}, \bm{z}^{1}_{m_2}, \ldots, \bm{z}^{T}_{m_M} = \mathcal{E}(\bm{x}^{1}_{m_1}, \bm{x}^{1}_{m_2}, \ldots, \bm{x}^{T}_{m_M}),
\end{equation}
where $\bm{x}^{t}_{m_j}$ is the token of the $m_j$-th modality at the time step $t$, $M$ denotes the number of input modality types, and $\bm{z}^{t}_{m_j}$ is the output token from the Transformer encoder $\mathcal{E}$.

\subsection{Hand Position and Ego-motion Predictors}
\label{sec: hand-ego}
We use two Transformer decoders, the hand decoder ($\mathcal{D}_\text{hand}$) and the ego-motion decoder ($\mathcal{D}_\text{ego}$), conditioned on the features from the encoder in an autoregressive manner.
Finally, the decoded feature for each future time step is fed into two MLP heads, $\mathcal{M}_\text{hand}$ and $\mathcal{M}_\text{ego}$, to predict the hand position and ego-motion for each time step.

\noindent\textbf{Transformer decoder.} 
For the hand Transformer decoder, the encoded left-hand token and the right-hand token of the last observation time $T$, $\bm{z}^T_{left}$ and $\bm{z}^T_{right}$, are used as the key and the value, and a learnable parameter is used as a hand learnable token $\bm{p}_\text{hand}$ for the query of the first forecasting time step ($\bm{q}^{T}_\text{hand}=\bm{p}_\text{hand}$):
\begin{equation}
\label{eq: 3}
    \bm{q}^{T+f}_\text{hand} = \mathcal{D}_\text{hand}(\bm{q}^{T}_\text{hand}, \ldots, \bm{q}^{T+f-1}_\text{hand}),
\end{equation}
where $\bm{q}^{T+f}_\text{hand}$ represents the decoded tokens for the future time step $T+f, f=\{1, \ldots, F\}$.
We perform the same operation for the ego-motion Transformer decoder.
The difference is the key, value, and query.
The key and value stem from the encoded ego-motion features at the last observed time step $T$, $\bm{z}^{T}_{ego}$, and the query is a learnable parameter for ego-motion $\bm{p}_\text{ego}$ ($=\bm{q}^{T}_\text{ego}$):
\begin{equation}
\label{eq: 5}
    \bm{q}^{T+f}_{ego} = \mathcal{D}_\text{ego}(\bm{q}^{T}_\text{ego}, \ldots, \bm{q}^{T+f-1}_\text{ego}).
\end{equation}

\noindent\textbf{MLP head.} 
We use multi-layer perceptrons (MLP), which take the decoded features from the Transformer decoder at each future time step for both hand position and ego-motion prediction.
$\mathcal{M}_\text{hand}$ predicts the coordinates of the left and right hands $\hat{\bm{h}}^{T+f}$ at the future time step $T+f$.
Similarly, $\mathcal{M}_\text{ego}$ predicts the nine elements of the homography matrix $\hat{\bm{e}}^{T+f}$:
\begin{equation}
\label{eq: 6}
    \hat{\bm{h}}^{T+f} = \mathcal{M}_\text{hand}(\bm{q}^{T+f}_\text{hand}),
\end{equation}
\begin{equation}
\label{eq: 7}
    \hat{\bm{e}}^{T+f} = \mathcal{M}_\text{ego}(\bm{q}^{T+f}_\text{ego}).
\end{equation}
Note that the weights of each MLP head ($\mathcal{M}_\text{hand}$ and $\mathcal{M}_\text{ego}$) are shared for each time step.

\subsection{Training Objective}
\label{sec: to}
In our training process, we use two types of losses: the hand forecasting loss $\mathcal{L}_\text{hand}$ and the ego-motion (nine elements of the homography matrix) estimation loss $\mathcal{L}_\text{ego}$.

\noindent \textbf{Hand forecasting loss.}
We adopt the self-adjusting smooth L1 loss, which was introduced in RetinaMask~\cite{retinamask}, as the objective function for hand forecasting:
\begin{equation}
\label{eq: 8}
    l_{i} =
    \begin{cases}
        0.5 w_{i} (h_{i} - \hat{h}_{i})^2 / \beta, & \lvert h_{i} - \hat{h}_{i} \rvert < \beta \\
        w_{i} (\lvert h_{i} - \hat{h}_{i}\rvert - 0.5 \beta), & \text{otherwise}
    \end{cases}\\
\end{equation}
\begin{equation}
    \mathcal{L}_\text{hand} = \frac{1}{4F} \sum_{i}^{} l_{i},
\end{equation}
where $h_{i}$ is a $i$-th element of a vector representing the $x,y$ ground truth coordinates of the left/right hands on $F$ future frames $\bm{h} \in \mathbb{R}^{4F}$, $\hat{\bm{h}}$ denotes predicted future hand coordinates, and $\beta$ is a control point that mitigates over-penalizing outliers.
If the hand is not observed in future frames, we pad $0$ into the $\hat{\bm{h}}$ and adopt a binary mask $\bm{w} \in \mathbb{R}^{4F}$ to prevent gradient propagation for these unobserved instances.

\noindent \textbf{Ego-motion estimation loss.}
We employ the L2 loss for ego-motion estimation loss:
\begin{equation}
\label{eq: 9}
    \mathcal{L}_\text{ego} = \frac{1}{9F} \sum_{i}^{} (e_{i} - \hat{e}_{i})^{2},
\end{equation}
where $\bm{e} \in \mathbb{R}^{9F}$ is a vector representing the elements of homography matrices on $F$ future frames.
$\mathcal{L}_\text{hand}$ and $\mathcal{L}_\text{ego}$ are linearly combined with a balancing hyperparameter $\alpha$ for the final training loss:
\begin{equation}
\label{eq: 10}
    \mathcal{L}_\text{total} = \mathcal{L}_\text{hand} + \alpha \mathcal{L}_\text{ego}.
\end{equation}
\section{Experiments}

\subsection{Datasets}
\noindent \textbf{EPIC-Kitchens 55}~\cite{epic-55}.
EPIC-Kitchens 55 is the dataset that only includes the daily activities videos in the kitchen.
It comprises a set of 432 egocentric videos recorded by 32 participants in their kitchens using a head-mounted camera.
We use the train/val split provided by RULSTM~\cite{rulstm}.

\noindent \textbf {Ego4D}~\cite{ego4d}.
The Ego4D dataset is the most recent large-scale egocentric video dataset. 
It contains 3,670 hours of egocentric videos of people performing diverse tasks, such as farming or cooking, and is collected by 931 people from 74 locations across nine different countries worldwide.
We follow the same train/val split protocol provided by Ego4D~\cite{ego4d}. 

Followed by the previous work~\cite{hoi-forecast}, we employ the egocentric hand-object detector~\cite{hod} with the same setup as the previous work and consider the center of detected hand bounding boxes as the ground truth hand positions for both left and right hands.

\subsection{Implementation Details}
\noindent \textbf{Experimental setup.}
We sample $T=8$ frames at 4 FPS (frames per second) as input observations and forecast 1 second with the future time step $F=4$ on both EPIC-Kitchens 55 and Ego4D.
We use the pre-trained ResNet-18~\cite{resnet} on ImageNet~\cite{imagenet} as the backbone to extract RGB and optical flow features. 
We adopt the hand and object detector from the egocentric video~\cite{hod} to detect left/right hand and object bounding boxes in each input frame, and FlowFormer~\cite{flowformer} is used to estimate the optical flow between consecutive frames.
We standardize RGB, optical flow, and ego-motion inputs using means and standard deviations of input modalities on the training dataset.
Note that the estimated homography matrices are normalized so that the element in the third row and the third column is one before standardization.

\noindent \textbf{Network architecture.}
We use the dimension size of a token $C=512$, $k=2$ for the top-$k$ confidence score with the threshold of $0.5$, and set the number of blocks in the encoder and decoder to $2$. 
Each block has $8$ attention heads in the encoder and decoder. 
Our MLPs for hand and ego-motion prediction consist of a linear layer, an activation function of ReLU~\cite{relu}, a Dropout~\cite{dropout} layer, and a final linear layer that outputs the hand positions and ego-motion at future frames.

\noindent \textbf{Optimization.}
We train the model for 30 epochs using the AdamW optimizer~\cite{adamw}, with a peak learning rate of $2e-4$, linearly increased for the first $5$ epochs of the training and decreased to $0.0$ until the end of training with cosine decay~\cite{cosine-decay}. 
We use weight decay of $1e-3$ and a batch size of $64$.
Regarding the parameters for the loss function, we empirically adapt the control point $\beta=5.0$ in \cref{eq: 8}, and the loss weight of $\alpha$, used in \cref{eq: 10}, is set to $1$.

\subsection{Evaluation Metrics}
The distance between the predicted and ground truth positions in 2D image space, measured in pixels, is used to evaluate future hand position prediction performance. 
Specifically, we adopt traditional metrics of trajectory prediction~\cite{sociallstm, socialgan, chang2019cvpr}: average displacement error and final displacement error.
Note that the metric is calculated using an image height scale of $256$ px.

\noindent\textbf{Average Displacement Error (ADE).} 
ADE is calculated as the $l_{2}$ distance between the predicted future hand positions and the ground truth positions in pixel averaged over the entire future time steps and both left and right hands.

\noindent\textbf{Final Displacement Error (FDE).} 
FDE measures the $l_{2}$ distance between the predicted future hand positions and ground truth positions at the last time step and is averaged over two hands.

\subsection{Comparison Methods}
We compare with the following methods:
\begin{itemize}
    \item \noindent\textbf{CVM}~\cite{cvm}.
    The Constant Velocity Model (CVM) is a simple but effective trajectory prediction method based on the assumption that the most recent relative motion is the most relevant predictor for the future trajectory.
    We compute the velocity $(v_x, v_y)$ between $t=T-1$ and $t=T$ for each hand (right, left), and future hand positions for $t=\{T+1,...,T+F\}$ are forecasted using $(v_x, v_y)$.
    \item \noindent\textbf{KF}~\cite{KF}.
    The Kalman Filter is an algorithm for estimating a dynamic system's state based on noisy measurements.
    It tracks the center of the bounding boxes of the hands with its scale and aspect ratio.
    Our implementation is based on the code provided by SORT~\cite{SORT}\footnote{\url{https://github.com/abewley/sort}}, which adopts a Kalman Filter to track the center of bounding boxes.
    \item \noindent\textbf{Seq2Seq}~\cite{seq2seq}.
    Seq2Seq employs Long Short-Term Memory (LSTM)~\cite{lstm} to encode temporal information in the observation sequence and decode the target location of the hands. 
    In our implementation, we adopt the embedding size of $512$, the hidden dimension of $256$, and the teacher forcing ratio of $0.5$ during training.
    \item \noindent\textbf{OCT}~\cite{hoi-forecast}.
    OCT simultaneously predicts contact points and the hand trajectory.
    It takes RGB features extracted by BNInception~\cite{BNInception}, bounding boxes of hands and objects, and their cropped visual features as input.
    We modified the model not to predict the contact point for a fair comparison.
    Our implementation of this model is based on the official implementation\footnote{\url{https://github.com/stevenlsw/hoi-forecast}}.
    \item \noindent\textbf{I3D + Regression}~\cite{ego4d}.
    This method is proposed as a benchmark for hand forecasting in the Ego4D dataset.
    The model is trained with the official hand forecasting code\footnote{\url{https://github.com/EGO4D/forecasting}}. 
\end{itemize}

The first two traditional approaches predict based only on past trajectories without training.
On the other hand, the last three methods above are recent advanced learning-based approaches in the hand forecasting task. 

\begin{table}[t]
\centering
\caption{\textbf{Intra-dataset evaluation.} We assess the performance of future hand forecasting on two large-scale egocentric datasets, Ego4D and EPIC-Kitchens 55. In terms of input modalities, the symbols $T_{h}, T_{o}, G_{r}, G_{f}, E$ represents \emph{trajectory information} of hands and objects, \emph{global information} of RGB and optical flow, and \emph{ego-motion information}, respectively. Note that no backbone is used in CVM, KF, and Seq2Seq as these methods predict based on past trajectories and do not input RGB or optical flow frames. The best values are shown in \textbf{bold}, and the second best values are shown with \underline{underline}.}
\centering
\scalebox{0.95}{
\begin{tabular}{lcccccc}
\toprule
\multirow{2.5}{*}{Method}   & 
\multirow{2.5}{*}{Input Modality}    & 
\multirow{2.5}{*}{Backbone} & 
\multicolumn{2}{c}{Ego4D} & 
\multicolumn{2}{c}{EPIC-Kitchens 55} \\
\cmidrule{4-7}
&&& ADE $\downarrow$ & FDE $\downarrow$ & ADE $\downarrow$ & FDE $\downarrow$\\
\midrule
CVM~\cite{cvm}                         & \scalebox{0.85}{$T_{h}$}      & - & 108.11 & 143.23 & 141.70 & 155.40\\ 
KF~\cite{KF}                           & \scalebox{0.85}{$T_{h}$}       & - & 71.23 & 72.87 & 70.58 & 75.60\\
Seq2Seq~\cite{seq2seq}                 & \scalebox{0.85}{$T_{h}$}       & - & 55.91 & 60.72 & 62.24 & 67.85\\
OCT~\cite{hoi-forecast}          & \scalebox{0.85}{$T_{h}, T_{o}, G_{r}$}     & BN-Inception & 49.40 & 54.73 & 53.85 & 59.06\\
I3D + Regression~\cite{ego4d}          & \scalebox{0.85}{$G_{r}$}       & 3D ResNet-50 & \underline{49.27} & \underline{53.04} & \underline{49.64} & \underline{54.83}\\
\rowcolor{lightgray}
Ours                                   & \scalebox{0.85}{$T_{h}, T_{o}, G_{r}, G_{f}, E$} & 2D ResNet-18 & \textbf{48.99} & \textbf{52.83} & \textbf{48.78} & \textbf{54.03}\\
\bottomrule
\end{tabular}}
\label{table: 1}
\end{table}

\begin{table}[t]
\centering
\caption{\textbf{Cross-dataset evaluation}. \textit{A} $\rightarrow$ \textit{B} in the first row indicates that the models are trained on the training set of dataset \textit{A} and tested on the validation set of dataset \textit{B}. We conduct two cross-dataset evaluations: (1) trained on EPIC-Kitchens 55 and evaluated on Ego4D and (2) trained on Ego4D and evaluated on EPIC-Kitchens 55.}
\centering
\scalebox{0.9}{
\begin{tabular}{lcccc}
\toprule
\multirow{2.5}{*}{Method}     & 
\multicolumn{2}{c}{EPIC $\rightarrow$ Ego4D} & 
\multicolumn{2}{c}{Ego4D $\rightarrow$ EPIC} \\
\cmidrule{2-5}
& ADE $\downarrow$ & FDE $\downarrow$ & ADE $\downarrow$ & FDE $\downarrow$ \\
\midrule
CVM~\cite{cvm}                & 108.11 & 143.23 & 141.70 & 155.40\\ 
KF~\cite{KF}                  & 71.23 & 72.87 & 70.58 & 75.60\\
Seq2Seq~\cite{seq2seq}        & 62.43 & 67.85 & 67.97 & 72.26\\
OCT~\cite{hoi-forecast} & \underline{57.74} & \underline{59.10} & 64.97 & 65.84\\
I3D + Regression~\cite{ego4d} & 59.72 & 61.72 & \underline{51.70} & \underline{58.37}\\
\rowcolor{lightgray}
Ours                          & \textbf{53.67} & \textbf{56.36} & \textbf{51.03} & \textbf{56.78}\\
\bottomrule
\end{tabular}}
\label{table: 2}
\end{table}

\subsection{Hand Forecasting Accuracy Comparison}
\noindent\textbf{Intra-dataset evaluation}.
We compare the performance of hand forecasting with the prior methods on two large-scale egocentric datasets.
\cref{table: 1} shows that the proposed method consistently outperforms the state-of-the-art methods.
Our proposed method surpasses OCT by $9.4\%$ (from $53.85$ to $48.78$) and I3D + Regression by $1.7\%$ (from $49.64$ to $48.78$) on the EPIC-Kitchens 55 dataset.
On the Ego4D dataset, our method exhibits similar performance on EPIC-Kitchens 55 and outperforms the prior works.
Furthermore, the poor performance of the constant velocity model~\cite{cvm}, which outperforms the learning-based approaches~\cite{Sadeghian2019CVPR,socialgan} for pedestrian trajectory prediction from exocentric videos, confirms that the 2D hand forecasting task from egocentric videos presents unique challenges due to ego-motion.

\begin{wrapfigure}{r}{0.5\linewidth}
\begin{center}
\vspace{-3em}
   \includegraphics[width=\linewidth]{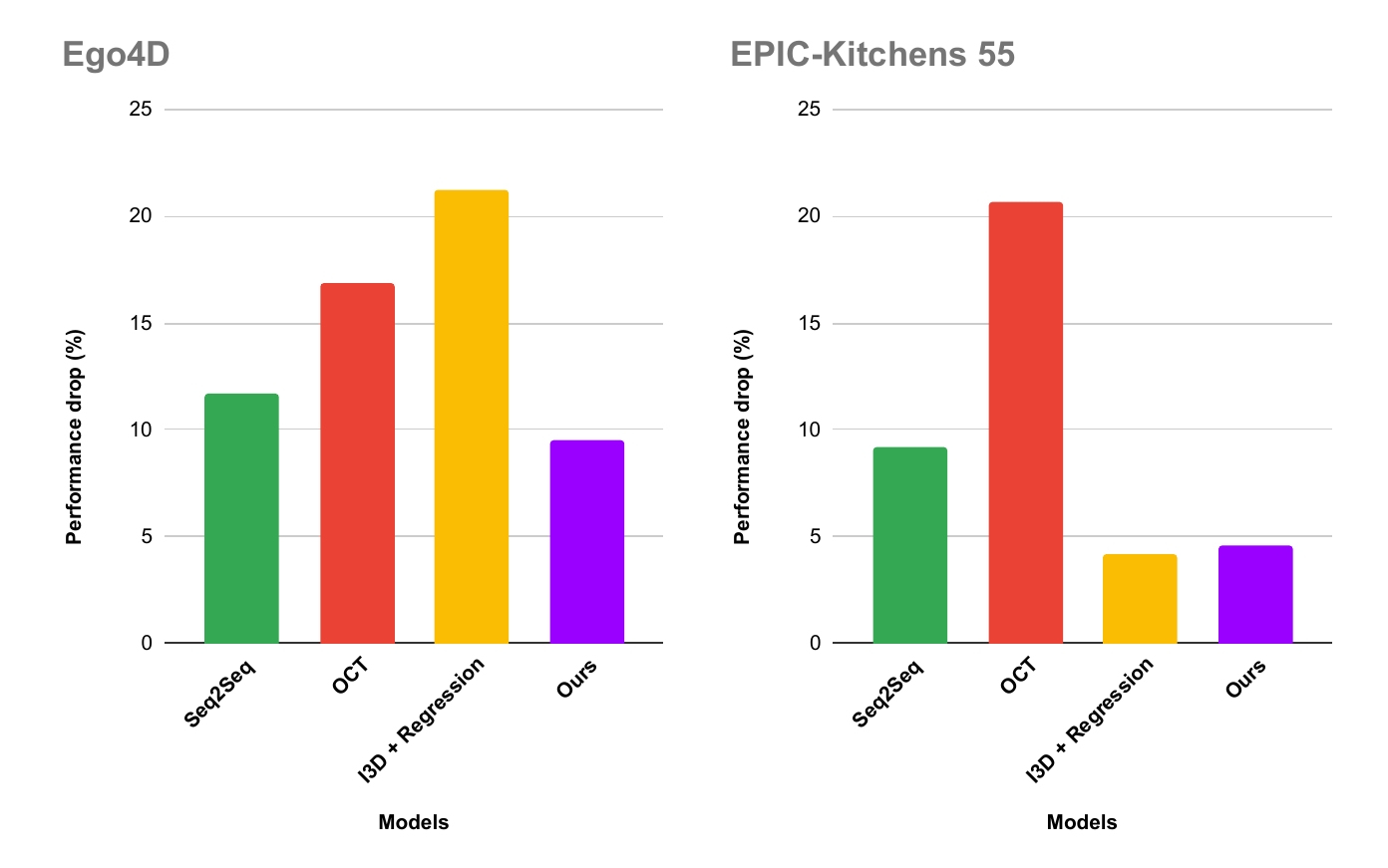}
\end{center}
\vspace{-2em}
   \caption{\textbf{The accuracy drop comparison.} The figure summarizes the accuracy drop percentage in the cross-dataset scenario from the accuracy in the intra-dataset scenario for each method. A lower value indicates that the performance does not drop by changing the scenario from intra-dataset to cross-dataset. We summarize the performance drop of the learning-based model as there is no performance degradation in non-learnable methods, such as CVM and KF.}
\label{fig: performance-drop}
\vspace{-2em}
\end{wrapfigure}

\noindent\textbf{Cross-dataset evaluation}.
We compare the generalization performance for future hand forecasting with the state-of-the-art methods in the cross-dataset scenario, where the domain of the test data is different from the training dataset.
\cref{table: 2} summarizes the generalization performance of the comparison methods and the proposed method.
Our proposed method surpasses OCT by $7.0\%$ on the Ego4D dataset, where the models are trained on the EPIC-Kitchens 55 dataset.

Moreover, the learning-based approaches (OCT, I3D+Regression, and Ours) demonstrate lower accuracy in the cross-dataset scenario as compared to intra-dataset evaluations (see \cref{fig: performance-drop}).
This performance decrease stems from dataset bias, as the two datasets originate from different distributions. 
The performance of I3D + Regression drops significantly ($21.1\%$) when the model is trained on EPIC-Kitchens 55 and tested on Ego4D.
On the other hand, although the accuracy is dropped in our method on Ego4D ($9.6\%$ dropping), the decrease is relatively small compared to other learning-based methods, thereby verifying the generalizability of the proposed method.

\begin{table}[t]
\centering
\caption{\textbf{Action category-level evaluation}. We compare the hand forecasting performance in the cross-dataset scenarios at the action category level with the conventional learning-based approaches. The results of five action categories, such as cooking, mechanic, arts/crafts, building, and gardening/farming, are summarized in the table.}
\centering
\scalebox{0.8}{
\begin{tabular}{l cccccccccc}
\toprule
\multirow{2.5}{*}{Method}     & 
\multicolumn{2}{c}{Cooking} & 
\multicolumn{2}{c}{Mechanic} & 
\multicolumn{2}{c}{Arts and crafts} & 
\multicolumn{2}{c}{Building} & 
\multicolumn{2}{c}{Gardening/farming} \\
\cmidrule(lr){2-11}
& 
ADE $\downarrow$ & FDE $\downarrow$ & 
ADE $\downarrow$ & FDE $\downarrow$ & 
ADE $\downarrow$ & FDE $\downarrow$ & 
ADE $\downarrow$ & FDE $\downarrow$ & 
ADE $\downarrow$ & FDE $\downarrow$ \\
\midrule
Seq2Seq~\cite{seq2seq}        & 58.45 & 60.73 & 59.78 & 62.83 & 64.60 & 66.85 & 68.28 & 70.11 & 64.42 & 66.52 \\
OCT~\cite{hoi-forecast} & 52.45 & 54.57 & \underline{53.63} & \underline{55.24} & \underline{62.52} & \underline{64.19} & \underline{63.06} & \underline{63.83} & \underline{57.49} & \underline{58.25}\\
I3D + Regression~\cite{ego4d} & \underline{48.26} & \underline{52.26} & 58.03 & 59.73 & 63.03 & 64.89 & 67.55 & 68.83 & 61.80 & 62.98 \\
\rowcolor{lightgray}
Ours                          & \textbf{47.32} & \textbf{51.33} & \textbf{47.53} & \textbf{51.02} & \textbf{58.89} & \textbf{61.28} & \textbf{59.83} & \textbf{62.30} & \textbf{53.16} & \textbf{55.68}\\
\bottomrule
\end{tabular}}
\label{table: category}
\end{table}

\noindent\textbf{Action category-level evaluation}.
We conduct action category-level evaluations in the cross-dataset scenario, where the models are trained on EPIC-Kitchens 55 and tested on each action category on Ego4D to assess the generalizability among unseen actions.
We focus on five major action categories on the Ego4D validation set: cooking, mechanic, arts/crafts, building, and gardening/farming.
\cref{table: category} demonstrates that our proposed method outperforms the prior learning-based methods across all categories.
This indicates that our proposed method is highly generalizable to unseen action categories.
In contrast, although the I3D + Regression method performs well in the cooking category, which is included in the training dataset, a significant performance gap can be seen in other categories compared to the cooking category.
This occurs because I3D + Regression tends to overfit to the context and background of the training data, particularly in the cooking category.

\subsection{Ablation Analysis}
\noindent\textbf{Input modality}.
The ablation study focuses on the input modalities to verify the contribution of each input component to the overall performance in intra/cross-dataset settings.
We experiment by removing each input modality: bounding boxes of objects, RGB frame, optical flow, and ego-motion information.
As shown in \cref{table: input-modality}, the absence of visual or flow information degrades the performance by $2.4\%$ (from $48.89$ to $50.08$) and $4.3\%$ (from $48.89$ to $51.00$) on intra-dataset evaluation on average, respectively.

Moreover, although the absence of object or ego-motion information outperforms the proposed method on intra-dataset evaluation, these methods degrade the prediction performance on cross-dataset scenarios.
This performance deterioration on cross-dataset scenarios indicates that leveraging all input modalities (the proposed method), including ego-motion information, is beneficial for unseen scenes.

\noindent\textbf{Loss}.
We also perform an ablation study on the loss function.
We evaluate the advantage of the ego-motion estimation loss term $\mathcal{L}_\text{ego}$ in \cref{eq: 10}.
\cref{table: loss component} shows that training the proposed method without the ego-motion estimation loss $\mathcal{L}_\text{ego}$ deteriorates hand forecasting performance by $1.6\%$ and $0.9\%$ in terms of ADE in the intra/cross-dataset scenario, respectively.
This degradation verifies the effectiveness of the proposed method, which forecasts the camera wearer's future ego-motion as an auxiliary task.

\begin{table}[t]
\begin{tabular}{cc}
\begin{minipage}[t]{0.54\textwidth}
\caption{\textbf{Input modality ablation study}. Ablation study on the input modalities on Ego4D and EPIC-Kitchens 55. 
We summarize the results of two scenarios, intra or cross-dataset. The last column is the result of the proposed method, which uses all the modal information.}
\centering
\scalebox{0.88}{
\begin{tabular}{cccccccc}
\toprule
\multirow{2.5}{*}{Object}    & 
\multirow{2.5}{*}{RGB}      & 
\multirow{2.5}{*}{Flow}      &
\multirow{2.5}{*}{Ego}       & 
\multicolumn{2}{c}{Intra}  & 
\multicolumn{2}{c}{Cross}  \\
\cmidrule(lr){5-8}
& & & & \scalebox{0.9}{ADE $\downarrow$} & \scalebox{0.9}{FDE $\downarrow$} & \scalebox{0.9}{ADE $\downarrow$} & \scalebox{0.9}{FDE $\downarrow$}\\
\midrule
             & $\checkmark$ & $\checkmark$ & $\checkmark$ & \underline{48.76} & 53.79 & \underline{52.78} & \underline{57.02}\\
$\checkmark$ &              & $\checkmark$ & $\checkmark$ & 50.08 & 54.83 & 53.30 & 57.54\\
$\checkmark$ & $\checkmark$ &              & $\checkmark$ & 51.00 & 54.78 & 54,74 & 57.93\\
$\checkmark$ & $\checkmark$ & $\checkmark$ &              & \textbf{48.35} & \textbf{53.24} & 52.89 & \underline{57.02}\\
\midrule
$\checkmark$ & $\checkmark$ & $\checkmark$ & $\checkmark$ & 48.89 & \underline{53.43} & \textbf{52.35} & \textbf{56.57}\\
\bottomrule
\end{tabular}}
\label{table: input-modality}
\end{minipage}
&
\hspace{0.02\textwidth}
\begin{minipage}[t]{0.42\textwidth}
\centering
\caption{\textbf{Loss component ablation study}. Ablation study on ego-motion estimation loss on the two datasets in intra and cross-dataset scenarios to verify the effectiveness of propagating ego-motion estimation loss.}
\scalebox{0.75}{
\begin{tabular}{l cccc}
\toprule
\multirow{2.5}{*}{Method}     & 
\multicolumn{2}{c}{Intra}  &
\multicolumn{2}{c}{Cross}  \\
\cmidrule(lr){2-5}
& \scalebox{0.99}{ADE $\downarrow$} & \scalebox{0.99}{FDE $\downarrow$} & \scalebox{0.99}{ADE $\downarrow$} & \scalebox{0.99}{FDE $\downarrow$}\\
\midrule
w/o $\mathcal{L}_\text{ego}$ & 49.66 & 54.26 & 52.84 & 57.08\\
w/ $\mathcal{L}_\text{ego}$ (Ours) & \textbf{48.89} & \textbf{53.43} & \textbf{52.35} & \textbf{56.57} \\
\bottomrule
\end{tabular}}
\label{table: loss component}
\end{minipage}
\end{tabular}
\end{table}

\begin{figure}[t]
\begin{center}
\includegraphics[width=0.88\linewidth]{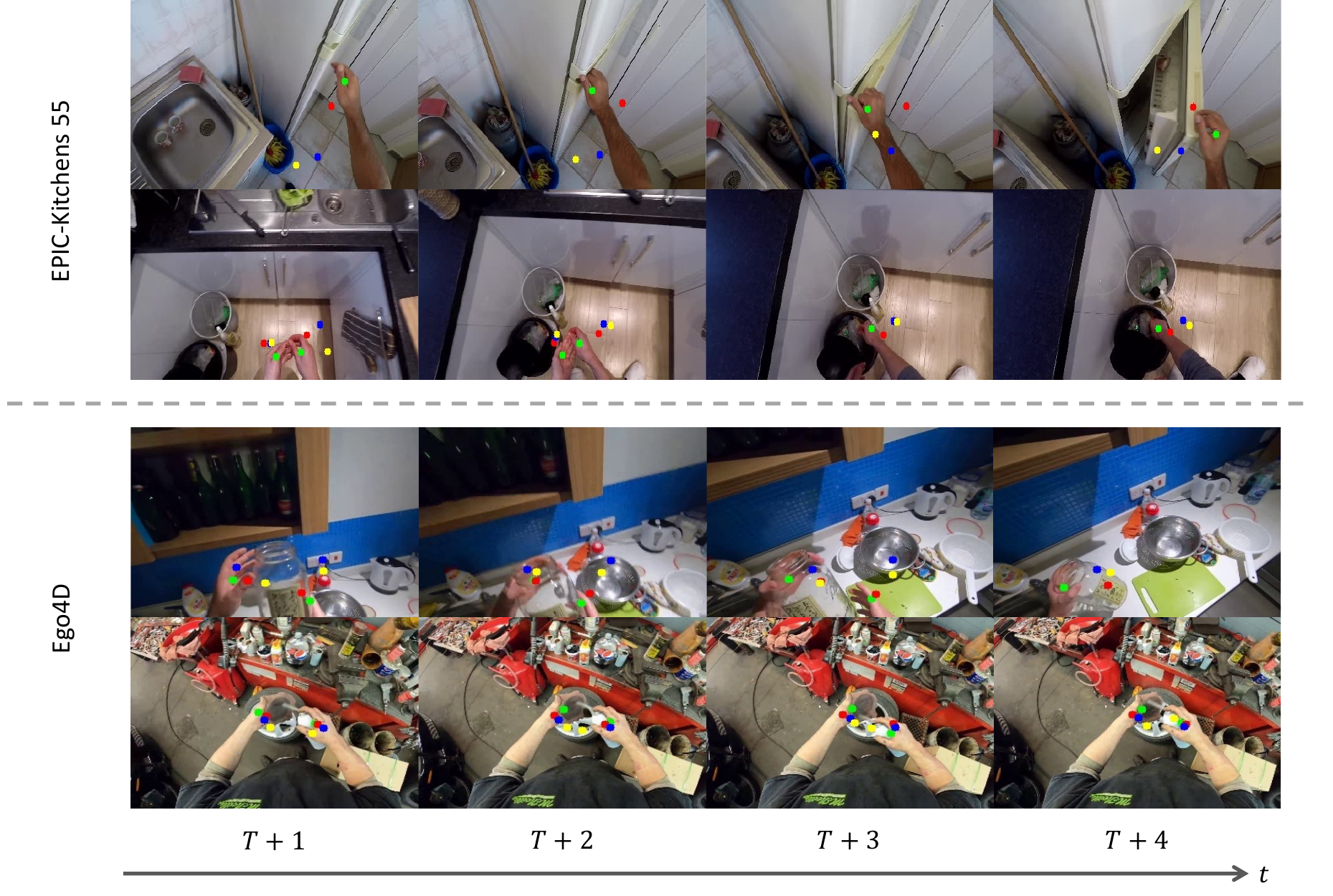}
\end{center}
   \caption{\textbf{Qualitative results}. We present two sequences of predictions each from Ego4D and EPIC-Kitchens 55. Dots colored in green, red, blue, and yellow represent the hand positions of the ground truth, the proposed method, I3D + Regression, and OCT, respectively.}
\label{fig: qualitative}
\end{figure}

\subsection{Qualitative Results}
The qualitative results on the Ego4D and EPIC-Kitchens 55 datasets are visualized in \cref{fig: qualitative}.
We present two sequences from EPIC-Kitchens 55 in the top two rows of the figure and two sequences from Ego4D in the bottom two rows.
In the second sequence from the top of EPIC-Kitchens, where the camera wearer turns left, the proposed method predicts the hand positions more accurately than the other methods.
This capability of prediction, even in the presence of ego-motion, verifies the effectiveness of our ego-motion-aware model.
\section{Conclusion}

\noindent\textbf{Conclusion}.
We present EMAG, the first model to explore the potential benefit of incorporating ego-motion into the hand forecasting task.
We propose leveraging the homography matrix to represent the camera wearer's ego-motion and to verify its effectiveness.
Furthermore, our proposed method utilizes multiple modalities to mitigate the susceptibility to overfitting to backgrounds or scene textures.
Experiments on two large-scale egocentric datasets, Ego4D and EPIC-Kitchens 55, demonstrate that our simple but effective approach outperforms the state-of-the-art hand forecasting methods in terms of accuracy and generalizability against unseen scenes and actions.

\noindent\textbf{Limitations and future work}.
Our proposed method leverages the trajectory information of hands and objects detected based on the off-the-shelf hand object detector~\cite{hod} from egocentric video.
Thus, the bias and errors from the off-the-shelf detector may still affect the input trajectory information.
In addition, the proposed method requires multiple pre-processing modules, such as hand object detection, optical flow estimation, and homography matrix estimation. 
However, efficient and real-time inference capabilities on edge devices are essential for forecasting in real-world applications.
We will leave this for our future efforts.

\section*{Acknowledgements}
This work was supported by JST BOOST, Japan Grant Number JPMJBS2409, and Amano Institute of Technology.

%
%
\bibliographystyle{splncs04}
\bibliography{egbib}

\clearpage

\title{EMAG: Ego-motion Aware and Generalizable \\ 2D Hand Forecasting from Egocentric Videos \\ \textmd{-- Supplementary Materials --}}

\titlerunning{EMAG Suppl.}

\author{Masashi Hatano\inst{1}\orcidlink{0009-0002-7090-6564} \and
Ryo Hachiuma\inst{2}\orcidlink{0000-0001-8274-3710} \and
Hideo Saito\inst{1}\orcidlink{0000-0002-2421-9862}}

\authorrunning{M.~Hatano et al.}

\institute{
Keio University \and
NVIDIA
}

\maketitle

\setcounter{figure}{4}
\setcounter{table}{5}
\begin{center}
    \centering
    \captionsetup{type=figure}
    \includegraphics[width=\linewidth]{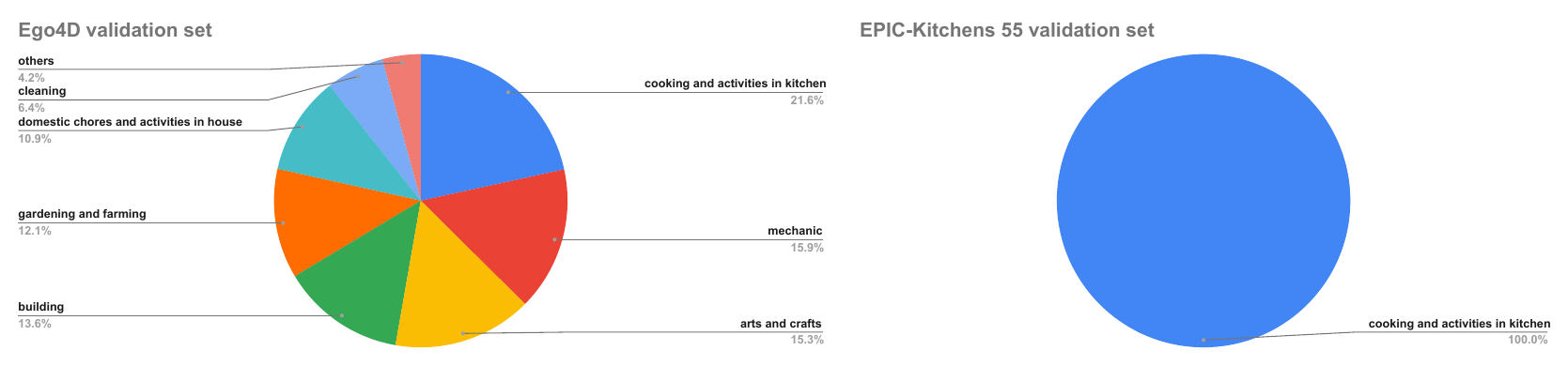}
    \captionof{figure}{\textbf{Scenario breakdown}. The left pie chart represents the scenario breakdown on the validation set of the Ego4D dataset. There are eight categories in total, including inside/outside scenes. The right pie chart represents the scenario breakdown on the validation set of the EPIC-Kitchens 55 dataset. The EPIC-Kitchens 55 dataset contains only one category, cooking and activities in the kitchen.}
    \label{fig: chart}
\end{center}

\thispagestyle{empty}
\appendix

\section{Dataset}
\subsection{Statistics}
This section provides statistics on two large-scale egocentric video datasets, Ego4D~\cite{ego4d} and EPIC-Kitchens 55~\cite{epic-55}.
\cref{fig: chart} presents pie charts illustrating the proportional distribution, categorized by action types or situations, of camera wearers within each validation set of the dataset.
The categories are summarized as follows:
\begin{itemize}
    \item \textbf{Cooking and activities in kitchen} contains videos where the camera wearer performs tasks in the kitchen, such as cutting vegetables, washing a pan, and putting dishes away on the shelf.
    \item \textbf{Mechanic} contains situations where the camera wearer uses specific mechanical tools to repair vehicles such as cars or bikes.
    \item \textbf{Arts and crafts} consist of indoor and outdoor scenarios, including activities such as painting and trimming excess materials.
    \item \textbf{Building} category contains a construction scene and a scene depicting brick fabrication.
    \item \textbf{Gardening and farming} consist of both small-scale and large-scale plant caring scenes.
    \item \textbf{Domestic chores and activities in house} contain activities in the house except for the situation in the kitchen, such as laundering, knitting, ironing, and playing cards.
    \item \textbf{Cleaning} category contains cleaning activities such as sweeping with a broom, mopping the floor, and washing a car.
    \item \textbf{Others} consist of various scenarios such as sports (playing basketball or working out at the gym), driving, walking a dog, and activities in the laboratory.
\end{itemize}
While all videos in the EPIC-Kitchens 55 dataset are categorized as cooking and activities in the kitchen, the Ego4D dataset contains various categories described above.
More than three-quarters of the videos in the validation set of Ego4D are composed of cooking and activities in the kitchen ($21.6\%$), mechanic ($15.9\%$), arts/crafts ($15.3\%$), building ($13.6\%$), and gardening/farming ($12.1\%$).

\begin{table}[t]
\caption{\textbf{Input modality ablation study}. Ablation study on the input modalities on Ego4D and EPIC-Kitchens 55. We evaluate the model in the intra and cross-dataset settings to verify the contribution of each input modality to the hand forecasting performance and the generalizability against novel scenes. In the last two rows, we summarize the results of two scenarios, intra and cross-dataset. The last column is the result of the proposed method, which uses all the modal information.}
\centering
\scalebox{0.78}{
\begin{tabular}{cccccccccccc|cccc}
\toprule
\multirow{2.5}{*}{Object}      & 
\multirow{2.5}{*}{RGB}      & 
\multirow{2.5}{*}{Flow}     &
\multirow{2.5}{*}{Ego}     & 
\multicolumn{2}{c}{\scalebox{0.8}{Ego4D $\rightarrow$ Ego4D}} & 
\multicolumn{2}{c}{\scalebox{0.8}{EPIC $\rightarrow$ Ego4D}} & 
\multicolumn{2}{c}{\scalebox{0.8}{EPIC $\rightarrow$ EPIC}} & 
\multicolumn{2}{c}{\scalebox{0.8}{Ego4D $\rightarrow$ EPIC}} &
\multicolumn{2}{|c}{\scalebox{0.8}{Intra}} & 
\multicolumn{2}{c}{\scalebox{0.8}{Cross}} \\
\cmidrule{5-16}
& & & & \scalebox{0.9}{ADE $\downarrow$} & \scalebox{0.9}{FDE $\downarrow$} & \scalebox{0.9}{ADE $\downarrow$} & \scalebox{0.9}{FDE $\downarrow$} & \scalebox{0.9}{ADE $\downarrow$} & \scalebox{0.9}{FDE $\downarrow$} & \scalebox{0.9}{ADE $\downarrow$} & \scalebox{0.9}{FDE $\downarrow$} & \scalebox{0.9}{ADE $\downarrow$} & \scalebox{0.9}{FDE $\downarrow$} & \scalebox{0.9}{ADE $\downarrow$} & \scalebox{0.9}{FDE $\downarrow$}\\
\midrule
             & $\checkmark$ & $\checkmark$ & $\checkmark$ & 49.02 & 53.00 & 54.25 & 56.79 & 48.50 & 54.57 & 51.31 & 57.25 & 48.76 & 53.79 & 52.78 & 57.02\\
$\checkmark$ &              & $\checkmark$ & $\checkmark$ & 51.02 & 54.30 & 54.09 & 57.15 & 49.14 & 55.35 & 52.90 & 57.93 & 50.08 & 54.83 & 53.30 & 57.54\\
$\checkmark$ & $\checkmark$ &              & $\checkmark$ & 50.82 & 53.77 & 55.57 & 57.70 & 51.17 & 55.78 & 53.90 & 58.15 & 51.00 & 54.78 & 54,74 & 57.93\\
$\checkmark$ & $\checkmark$ & $\checkmark$ &              & 49.04 & \textbf{52.69} & 54.22 & 57.01 & \textbf{47.66} & \textbf{53.79} & 51.55 & 57.02 & \textbf{48.35} & \textbf{53.24} & 52.89 & 57.02\\
\midrule
$\checkmark$ & $\checkmark$ & $\checkmark$ & $\checkmark$ & \textbf{48.99} & 52.83 & \textbf{53.67} & \textbf{56.36} & 48.78 & 54.03 & \textbf{51.03} & \textbf{56.78} & 48.89 & 53.43 & \textbf{52.35} & \textbf{56.57}\\
\bottomrule
\end{tabular}}
\label{table: input-modality-all}
\end{table}

\section{Further Results}
\subsection{Input Modality Ablation}
\cref{table: input-modality-all} shows all four intra/cross-dataset scenarios using two datasets, trained and evaluated on either Ego4D or EPIC-Kitchens 55, and the aggregated results for intra/cross-dataset scenarios.

\begin{table}[t]
\caption{\textbf{Loss component ablation study}. Ablation study on ego-motion estimation loss on two datasets in intra and cross-dataset scenarios to verify the effectiveness of estimating future ego-motion as an auxiliary task.}
\centering
\scalebox{0.9}{
\begin{tabular}{lcccccccc}
\toprule
\multirow{2.5}{*}{Method}     & 
\multicolumn{2}{c}{Ego4D $\rightarrow$ Ego4D} & 
\multicolumn{2}{c}{EPIC $\rightarrow$ Ego4D}  &
\multicolumn{2}{c}{EPIC $\rightarrow$ EPIC}  &
\multicolumn{2}{c}{Ego4D $\rightarrow$ EPIC} \\
\cmidrule(lr){2-9}
& \scalebox{0.99}{ADE $\downarrow$} & \scalebox{0.99}{FDE $\downarrow$} & \scalebox{0.99}{ADE $\downarrow$} & \scalebox{0.99}{FDE $\downarrow$} & \scalebox{0.99}{ADE $\downarrow$} & \scalebox{0.99}{FDE $\downarrow$} & \scalebox{0.99}{ADE $\downarrow$} & \scalebox{0.99}{FDE $\downarrow$} \\
\midrule
w/o $\mathcal{L}_\text{ego}$ & 49.59 & 53.15 & 53.85 & 56.57 & 49.72 & 55.37 & 51.83 & 57.59 \\ 
w/ $\mathcal{L}_\text{ego}$ (Ours) & \textbf{48.99} & \textbf{52.83} & \textbf{53.67} & \textbf{56.36} & \textbf{48.78} & \textbf{54.03} & \textbf{51.03} & \textbf{56.78} \\
\bottomrule
\end{tabular}}
\label{table: loss}
\end{table}

\noindent \textbf{Analysis}.
As shown in \cref{table: input-modality-all}, our proposed method is outperformed by the model that omits object or ego-motion information in the scenario, where models are trained and tested on EPIC-Kitchens 55.
This occurs due to the overfit to the context of the cooking category.
Methods lacking object or ego-motion information tend to rely more on RGB information to predict future hand positions than the proposed method that leverages all modalities.

\begin{wrapfigure}{r}{0.55\linewidth}
\begin{center}
\vspace{-3.5em}
\includegraphics[width=\linewidth]{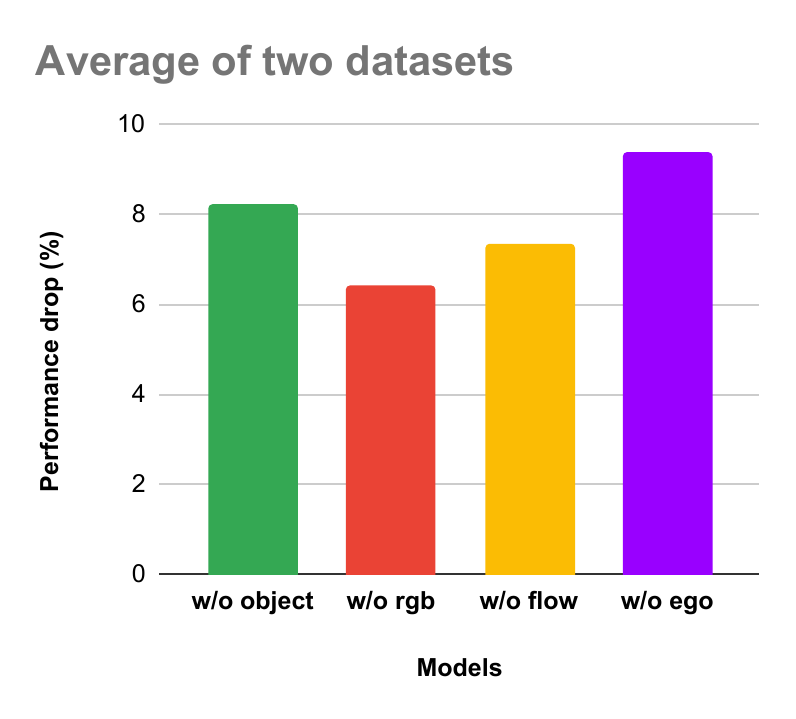}
\end{center}
\vspace{-2em}
   \caption{\textbf{The performance drop of each model that lacks one of the input modalities}. The green, red, yellow, and purple bar charts represent models without objects, RGB, optical flow, and ego-motion information.}
\label{fig: pd_input_modality}
\vspace{-3em}
\end{wrapfigure}

\noindent \textbf{Generalizability of each input modality}.
We further analyze the generalizability of each input modality: the trajectory of bounding boxes of objects, RGB, optical flow, and ego-motion information.
\cref{fig: pd_input_modality} shows the drop in performances for each model that is missing one of the four input modalities, from the intra-dataset scenario to the cross-dataset scenario in the average of two datasets.
The smaller the performance drop is, the more the leveraged modalities (the other three modalities other than the lacking modality) contribute to the generalizability against unseen data.
The performance drops of the method without object, RGB, optical flow, and ego-motion, are $8.24\%$, $6.43\%$, $7.33\%$, and $9.39\%$, respectively.
This confirms that RGB is the most susceptible to unseen data, as RGB depends on appearance, which leads to the overfit to backgrounds or the contexts, and the ego-motion information (homography) is the most generalizable input modality among the four modalities against novel scenes.

\subsection{Loss Component Ablation}
\cref{table: loss} shows the hand forecasting performance of whether adopting the ego-motion estimation loss $\mathcal{L}_\text{ego}$ in all four intra/cross-dataset scenarios.
The method without using $\mathcal{L}_\text{ego}$ deteriorates the performance in all intra/cross-dataset scenarios, verifying the effectiveness of estimating future ego-motion as an auxiliary task for both intra and cross-dataset settings.

\begin{table}[tb]
\small
\caption{Ablation study of ego-motion representation. }
\centering
\scalebox{1}{
\begin{tabular}{lcc}
\toprule
\multirow{2.5}{*}{Method}  & 
\multicolumn{2}{c}{Ego4D$\rightarrow$EPIC} \\ 
\cmidrule{2-3}
& ADE $\downarrow$& FDE $\downarrow$\\
\midrule
Background flow      & 52.08 & 58.03 \\
Ours    & \textbf{51.03} & \textbf{56.78} \\
\bottomrule
\end{tabular}}
\label{tab:ablation}
\end{table}

\subsection{Ego-motion Representation}
We conducted an additional ablation study on ego-motion representation, considering the homography matrix and background optical flow (\cref{tab:ablation}). The proposed homography matrix representation outperformed the background optical flow representation in cross-scenarios, underscoring the effectiveness of the proposed ego-motion representation.


\end{document}